\documentclass[letterpaper]{article} % DO NOT CHANGE THIS
\usepackage{aaai23}  % DO NOT CHANGE THIS
\usepackage{times}  % DO NOT CHANGE THIS
\usepackage{helvet}  % DO NOT CHANGE THIS
\usepackage{courier}  % DO NOT CHANGE THIS
\usepackage[hyphens]{url}  % DO NOT CHANGE THIS
\usepackage{graphicx} % DO NOT CHANGE THIS
\usepackage{natbib}  % DO NOT CHANGE THIS AND DO NOT ADD ANY OPTIONS TO IT
\usepackage{caption} % DO NOT CHANGE THIS AND DO NOT ADD ANY OPTIONS TO IT
\usepackage{algorithm}
\usepackage{xcolor}
\usepackage{tikz}
\usepackage{tikz-cd}
\usepackage{tikzit}
\usepackage{amsmath,amsfonts,amssymb,amsthm}
\usepackage{booktabs}
\usepackage{listings}
\usepackage{enumitem}
\usepackage{mathtools}
\usepackage{inconsolata}
\usepackage{algpseudocode}
\usepackage{makecell}
\usepackage{url}
\usepackage{anyfontsize}

\newcommand{\cat}[1]{\mathsf{#1}}
\newcommand{\Set}{\mathsf{Set}}
\newcommand{\catSet}[1]{\cat{#1}\text{-}\Set}

\tikzstyle{morphism}=[fill=white, draw=black, shape=rectangle, minimum width=1cm, minimum height=0.75cm]

\tikzstyle{functor}=[->, >=stealth]
\tikzstyle{dotted}=[-, dashed]
\tikzstyle{double arrow}=[<->, >=stealth]
\theoremstyle{definition}
\newtheorem{definition}{Definition}[section]

\usepackage{newfloat}
\usepackage{listings}
\DeclareCaptionStyle{ruled}{labelfont=normalfont,labelsep=colon,strut=off} % DO NOT CHANGE THIS
\floatstyle{ruled}
\newfloat{listing}{tb}{lst}{}
\floatname{listing}{Listing}
\title{A Categorical Representation Language and Computational System for Knowledge-Based Robotic Task Planning}
\author {
    Angeline Aguinaldo\textsuperscript{\rm 1,\rm 2},
    Evan Patterson\textsuperscript{\rm 3},
    James Fairbanks\textsuperscript{\rm 4},
    William Regli\textsuperscript{\rm 1},
    Jaime Ruiz\textsuperscript{\rm 4}
}
\affiliations {
    \textsuperscript{\rm 1} University of Maryland, College Park\\
    \textsuperscript{\rm 2} Johns Hopkins University Applied Physics Laboratory\\
    \textsuperscript{\rm 3} Topos Institute\\
    \textsuperscript{\rm 4} University of Florida\\
    \{aaguinal, regli\}@cs.umd.edu, 
    evan@topos.institute,
    \{fairbanksj, jaime.ruiz\}@ufl.edu
}

\usepackage{bibentry}
\begin{document}

\maketitle

\begin{abstract}
Classical planning representation languages based on first-order logic have preliminarily been used to model and solve robotic task planning problems. Wider adoption of these representation languages, however, is hindered by the limitations present when managing implicit world changes with concise action models. To address this problem, we propose an alternative approach to representing and managing updates to world states during planning. Based on the category-theoretic concepts of $\cat{C}$-sets and double-pushout rewriting (DPO), our proposed representation can effectively handle structured knowledge about world states that support domain abstractions at all levels. It formalizes the semantics of predicates according to a user-provided ontology and preserves the semantics when transitioning between world states. This method provides a formal semantics for using knowledge graphs and relational databases to model world states and updates in planning. In this paper, we conceptually compare our category-theoretic representation with the classical planning representation. We show that our proposed representation has advantages over the classical representation in terms of handling implicit preconditions and effects, and provides a more structured framework in which to model and solve planning problems.
\end{abstract}

\section{Introduction}

% \subsection{Open problems in planning applications}

% \subsubsection{Using structured knowledge for planning}

% \subsubsection{Online planning}

In robotic task planning, tracking all the implicit effects and relationships in the world state can be very challenging, especially when working in complex environments. As a result, planning systems often rely on heuristics and simplifying assumptions, which can lead to suboptimal or even incorrect solutions \cite{Wilkins2001,Gil1990}. To address these challenges, researchers have sought to combine techniques from the fields of knowledge representation and automated planning which has formed the subfield called \textit{knowledge-based planning} \cite{Wilkins2001}. The goal of this research is to develop more structured and efficient representations of the world state that can capture the complex relationships and dependencies that arise in real-world planning problems. For example, some approaches use structured knowledge representations, such as ontologies \cite{Kang2009,Beetz2018,Tenorth2017} and knowledge graphs \cite{Amiri2022,Galindo2008,Miao2023,Agia2022}, to model the world state. It is natural to conclude that knowledge-based planning, as it stands, may be better suited to solve the problem of tracking implicit preconditions and effects. On the contrary, knowledge-based planning has not been widely adopted because it requires translating these rich and complex representations into propositional facts so that they can be interpreted using a classical planning languages like PDDL \cite{Knoblock1998}. Because of this, they become computationally intensive search problems in the absence of search space reduction methods.

% As a result, knowledge-based planning is met with the challenge of finding the right balance between abstraction and representation that is both comprehensive and computationally efficient. Domain abstractions should ideally simplify world state and action models without losing essential characteristics. Representations, on the other hand, influence these domain abstractions according to their expressivity and structure. In particular, they dictate how to express criteria that must be met for actions to be taken and how the state will be updated. An ideal knowledge-based planning representation should be able to preserve the semantics of the domain when updating states. Formal languages like propositional and first-order logic are often used to define these representations \cite{Russell2021}; however, they lack the structure to express more complex relationships such as hierarchy and composition making them ill-equipped to solve the problems faced by knowledge-based planning representations. 

To address these concerns, we propose a world state representation based on the category-theoretic concepts of $\cat{C}$-sets \cite{Patterson2021} and double-pushout (DPO) rewriting \cite{Brown2022}. Our representation not only manages structured knowledge about the world state, but also formalizes the semantics of predicates according to a user-provided ontology, ensuring that the semantics of the world state are preserved and implicit preconditions and effects are handled when transitioning between states.

\section{Related Work}
\label{sec:related}

% Only recently have description logics been used to provide more structure to the world states in planning.

% \subsection{Extensions of representation languages for planning}

% Formalisms based on restricted first-order logic (FOL) for planning are limited in their ability to accurately describe real-world domains. Various syntactic and semantic extensions have been proposed to increase the interpretability and expressibility of the language. These extensions include typing of variables and predicates, which has proven to be particularly effective. PDDL 1.2 \cite{Knoblock1998} introduced support for typing objects in the domain, which enables the creation of a type hierarchy or ontology by defining a new abstract type. For instance, by asserting \texttt{(:type location building)}, we can consider \texttt{location} to be the abstract or parent type of \texttt{building}. To ensure that predicate arguments accept all constants of the parent type, we should use the type \texttt{location} rather than \texttt{building}, as in the expression \texttt{in(?x - location)}. Note that this specification does not capture the hierarchical relationship between \texttt{location} and \texttt{building}. During planning, the typing is used to validate that the constants used in the action operator adhere to the type mapping. However, inferences regarding how constants relate to an upper ontology are solely dependent on the domain designer's specification and cannot be inferred using the existing formal structures.

% DKEL \cite{Haslum2003}

% \subsection{Robotic task planning using scene graphs}

We foresee the nearest application of our approach as being robotic task planning using scene graphs. Scene graphs are a specialized version of knowledge graphs that restrict its elements, attributes, and relations to facts obtained through vision-based perception and inference \cite{Chang2023}. In scene graphs, ontologies are often used to align scene data to class hierarchies. In planning, these ontologies can be used to enrich facts in the planning domain; however, these ontologies are often integrated in the planning decisions in ad hoc ways. For example, Galindo et al present a two-part knowledge representation system, which includes (i) spatial information about the robot environment in the form of a scene graph, and (ii) an ontology that describes the hierarchical relationships between concepts \cite{Galindo2008}. A function mapping elements in the scene graph to concepts in the ontology is defined. Both facts obtained through the scene graph and facts obtained from the ontology are translated into propositions in the domain. This approach results in an explosion of facts that requires a pruning step in order to be tractable for classical planning approaches. 

This problem of scene graphs being too complex, with numerous vertices and edges, is long standing. To mitigate this, planners can employ procedures that determine which attributes of the scene are most relevant while also preserving the semantics provided by the class hierarchy and item features. An example of a planner designed for this purpose is the SCRUB planner \cite{Agia2022}. SCRUB is a planner-agnostic procedure that prunes the state space to include only the relevant facts within a scene graph. It is paired with SEEK \cite{Agia2022}, a planner-agnostic procedure that scores items in the scene based on an importance score produced by a graph neural network \cite{Silver2021}. All elements that are ancestors to the relevant elements, according to some threshold, are preserved as facts in the state. Both the SCRUB and SEEK procedures dramatically reduce the number of facts needed to characterize the world state. The facts in the world are translated into binary predicates and passed to a classical planner. This provides a heuristic-based measure for identifying relevant facts which can, like most heuristics, produce inaccurate approximations. The combinatorial approach we propose uses the existing semantic structure to determine relevance.

Miao et al take an alternate approach to using scene graphs to describe world states and action models \cite{Miao2023}. In their approach, action operators are specified in terms of an initial state subgraph, a final state subgraph, and an intermediate subgraph. For each element and relation in these subgraphs, the global scene graph is updated by adding elements into the scene graph that are discussed in the action model. If elements are referenced in the final state subgraph that are not present in the scene, they are introduced as isolated vertices. These vertices are connected to the global scene graph by consulting an external knowledge base that contains a type hierarchy. Their procedure searches for a matching element type, identifies a parent type that exists in the scene graph, and defines an edge from that element to the existing type. This is a fragile approach to resolving changes in the world state because it relies on an external knowledge base to be correctly and completely instantiated in order for new information to be properly integrated.

Overall, using scene graphs as a representation for world states during planning has gained attention. These methods satisfy the need to support rich and complex representations of the world using ontologies but still suffer in their ability to integrate ontological information in a way that does not cause a state explosion. In our approach, the ontology (schema) is an integral part of the formalism and with it comes specialized tooling for manipulating such data. 

% \subsection{Answer Set Programming}
% An alternative language has been proposed for planning called answer set programming (ASP) \cite{Lifschitz2002} that is designed for knowledge-intensive reasoning. ASP has a nonmonotonic property that allows a planner to remove information from the state when new information becomes available. ASP-based planning identifies a program which is an unordered set of rules that when applied to the initial state achieves the goal and every state is called an answer set which is a set of propositional facts including negation.

\section{Preliminaries}
\label{sec:prelim}

% The purpose of identifying a compact and ergonomic world state representation for planning is to prevent the need to account for all aspects of the world and how it changes when different actions are applied. The representation should be capable of deriving the implicit characteristics of the world in light of explicit ones. In the vernacular of logical representations, we can translate "implicit characteristics" to mean inferences about logical entailments between sentences.

% When deciding on a world state representation for planning, it is useful to consider representations as languages that supply the syntax and semantics used to discuss truthiness of the world and relations from one world state to another. A more structured language may provide more syntax and semantics that impose constraints on valid truth constructions and world changes, whereas a less structured language may provide fewer of each.

In this section, we explain the mathematical model for planning as a state transition system and briefly discuss the use of category theory.

\subsection{Planning as a State Transition System}

% The prevalent classical planning language for describing domains and problems is the Problem Domain Definition Language (PDDL) \cite{Ghallab1998}. The schema adopted by PDDL at inception was based on the language used by the Stanford Research Institute Problem Solver (STRIPS) \cite{Fikes1971}. This requires a set of propositions, $F$, a set of operator schemas with preconditions and effects, $O$, an initial state, $I$, and a goal state, $G$,  and operates under the \textit{closed world assumption}---all absent information is negative information. In other words, a STRIPS-based PDDL planning model can be defined as $P = <F, O, I, G>$. State models, $S(F)$, is a set of world states, $s \in S(F)$, such that its elements are propositions from $F$. $A$ is the ground set of actions obtained from $O$. A transition function, $f: A \times S \rightarrow S$, maps between states according to the action applied. A corresponding cost is computed using a cost function, $\sigma: A \times S \rightarrow \mathbb{R}$. A plan, $\pi = <a_{i}, a_{i+1}, ... a_{n}>, a_{i} \in A$, is the sequence of actions that transition from the initial state, $s_0 = I$, to the goal state, $s_G = G$, according to the transition function, $f(a, s)$ and cost, $c(\pi) = \sum_{j=i}^{n} \sigma(a_j, s_j)$  \cite{Geffner2013}. There have been a number of solver heuristics and search algorithms developed that design cost functions according to soundness and optimality in order to identify plans efficiently \cite{Ghallab2004}.

The state transition system model of planning is a formal representation of a planning problem that describes the state space of the problem and the possible actions that can be taken to move between states. In this model \cite{Ghallab2004}, a \textit{planning problem} can be defined as a tuple $P = \langle S, A, \gamma \rangle$, where:

\begin{itemize}
\item The \textit{state space} $S = \{s_0, s_1, s_2, ...\}$ is the set of all possible states. A state $s \in S$ represents a snapshot of the world at a particular point in time. It ideally includes all relevant information about the state of the world, such as the location of items and their properties.
\item The \textit{action space} $A = \{a_0, a_1, a_2, ...\}$ is the set of all possible actions. An action represents a transition from one state to another state. 
% Each action $a \in A$ has preconditions $\operatorname{Pre}(a)$, which are the conditions that must be satisfied for the action to be applicable, and effects $\operatorname{Eff}(a)$, which describe the changes that the action makes to the world. $\operatorname{Pre}(a)$ and $\operatorname{Eff}(a)$ can be expressed as logical formulas or sets of propositions which are discussed in Section $\ref{sec:classical}.
\item The \textit{transition function} $\gamma: A \times S \rightarrow S$ is a partial function that, where it is defined, maps an action and a state to the next state.
\end{itemize}

An action $a \in A$ is \textit{applicable} at state $s \in S$ if $\gamma(a, s)$ is defined. A \textit{plan}, $\pi = \langle a_1, a_2, \dots, a_{n} \rangle, a_{i} \in A$, is any sequence of actions. It is a \textit{solution} to the planning problem if it transitions from the initial state $s_0$ to the goal state $s_g$, i.e., $\gamma(a_i, s_{i-1}) = s_i$ for $i = 1,\dots,n$ and $s_n = s_g$.

The state transition system model of planning provides a structured and formal way to represent planning problems and reason about the possible sequences of actions that can be taken to achieve a goal. The manner in which states, actions, and transition functions are represented both classically and categorically is the focus of this paper.

\subsection{Category Theory}

Category theory is a mathematical framework that aids in understanding the relationships between different mathematical structures. In recent years, it has found important applications in systems engineering and design \cite{Breiner2019a,Breiner2016,Censi2015}, robotics and planning \cite{Aguinaldo2021,Aguinaldo2021a,Master2019}, and physics \cite{Baez2011,Abramsky2008}. For the purposes of planning, category theory provides an alternative mathematical language for representing world states and action models. 

\section{Classical Representations}
\label{sec:classical}

The literature on classical and neoclassical planning encompasses a variety of approaches to modeling the state of the world. Among these, the most widely employed domain-independent representations are the set-theoretic, classical, and state-variable representations \cite{Ghallab2004}.

\subsubsection{Existing Representations for World States, $S$}

The classical representation of world states is based on a restricted form of first-order logic in which the world states are represented as a conjunction of \textit{literals}, where a literal is an atomic proposition or its negation. These literals can encode facts such as the location of items, the state of various sensors, and other relevant information. Logical operators such as AND, OR, and NOT are used to combine these literals into more complex expressions that can be used to describe the relationships between different parts of the world state. World states can be lifted to abstract states consisting of a conjunction of \textit{predicates}, where predicates are $n$-ary relations between variable symbols. For example, the predicate $\mathrm{on}(x, y)$ might represent the fact that something represented by $x$ is on top of another thing represented by $y$ or, equally plausible, the converse. It is important to note that the semantics of the predicates must be established informally, say through external documentation or word of mouth. A predicate becomes a grounded literal when the variable symbols, like $x$ and $y$, are assigned to symbols that are constant. For instance, the grounded literal $\mathrm{on}(\mathrm{b_1}, \mathrm{b_2})$ could represent a block called $\mathrm{b_1}$ that is on top of a block called $\mathrm{b_2}$.

\subsubsection{Existing Representations for Actions, $A$}

The classical representation defines action schemas in terms of preconditions, effects, and parameters. Preconditions describe the conditions that must be true for the action to be applicable, while effects describe the changes that the action makes to the state of the world. Preconditions and effects are defined in terms of predicates where their variable symbols, also known as parameters, can take on different values depending on the context in which the action is used. The classical planning representation assumes that actions are deterministic, meaning that their effects are completely specified and do not depend on any probabilistic factors. Additionally, the classical representation assumes that actions can be arranged in any order, allowing for a wide variety of possible plans to be generated. In practice these action schemas can be augmented with types; however, types are not a formal feature of the logical system.

\subsubsection{Applicability}
\label{sec:applicability}

In classical planning, the applicability of actions to a world state is determined by a set of preconditions specified for each action. In the classical representation, preconditions are represented as conjunctions of literals. To check whether an action is applicable in a particular world state, the planner examines whether the preconditions of the action are satisfied by the current state. To do this, a planner typically uses logical inference techniques within propositional logic, first-order logic, or other logical formalisms. Once grounded, conjunction of predicates can be represented as sets of ground literals. The planner can then check whether the precondition is a subset of the world state, which means that the preconditions are satisfied in the world state. It is possible to model domain-specific implications, called derived predicates in PDDL \cite{Thiebaux2005}, to incorporate implicit preconditions when determining applicability of an action; however, derived predicates do not make any changes to the world state.

% \textit{Set-based Inclusion}

% \textit{Logical Entailment}
% \label{sec:entail}
% In first-order logic, logical entailment discusses the type of inclusion of a given formula, $f$, in a knowledge base, $K$. 

% \begin{definition}[Entailment ($\models$)]
%   Let $f$ be an FOL expression and $K$ is a set of FOL expressions. A set of models for $f$, $\mathcal{M}(f)$, are all the set of all binary assignments for the propositional symbols contained in $f$ that cause $f$ to be true. $\mathcal{M}(K)$ refers to intersection of the set of models for all formulas in $K$. $K \models f \iff \mathcal{M}(K) \cap \mathcal{M}(f) = \mathcal{M}(K)$. In the language of set theory, $K$ entails $f$ can be seen as $K$ is a subset of $f$.
% \end{definition}

% Entailment of formulas is determined by unification and resolution methods \cite{}. 
% In planning, applicability can be thought of as a type of entailment.

\subsubsection{Transition, $\gamma$}

A state transition occurs when an action is applied to a world state, yielding a new world state. Similar to the preconditions, the world state can be interpreted as a set of literals. Therefore, when an action is applied, the new state is identified by applying set operations to the set representing the current world state. Positive effects add ground literals to the set and negative effects, those prefixed with NOT, subtract ground literals from the set to produce the new world state.

\section{Categorical Representation}
\label{sec:categorical}

To effectively manage world states during planning and plan execution, we propose to adopt categorical logic as a logical formalism. An essential feature of categorical logic is the ability to capture the relationship between logical syntax (theories) and semantics (models) using functors, through a paradigm known as \textit{functorial semantics}. This differs from the standard formalism of first order logic because it gives the syntax of the language status as an algebraic object independent of its semantics. 

\subsection{World States as $\cat{C}$-sets}

In our approach, we assume world states in robot task planning domains can be effectively modeled using scene graphs, which are structured representations of the elements, attributes, and relations based on a domain-specific ontology. Recall that scene graphs are a specialized form of knowledge graphs that focus on the visual aspects of a scene, with nodes representing items and edges representing the spatial relationships between them \cite{Chang2023}. These graphs can be used to capture a wide range of information about a world state, including the physical and semantic properties of items, their interactions with each other, and the context in which they exist.

Our proposed representation provides a sort of denotational semantics for scene graphs using the category $\catSet{C}$. Let $\cat{C}$ denote a small category, called a \textit{schema}. A \textit{$\cat{C}$-set}\footnote{A \textit{$\cat{C}$-set} also known as a \textit{copresheaf on $\cat{C}$}.} is a functor\footnote{We direct the unfamiliar reader to \cite{Leinster2016} for the definitions of categories, functors, natural transformations, limits, and colimits.} from $\cat{C}$ to the category $\Set$. The schema is a category whose objects we interpret as types and whose morphisms describe ``is-a'' and other functional relationships between types. The category $\Set$ is the category of sets and functions. Thus, a $\cat{C}$-set is a functor that sends types to sets and type relationships to functions. On this interpretation, $\cat{C}$-sets are a simple but useful model of relational databases \cite{Spivak2011}.
% and have a full-featured implementation in [Catlab.jl](https://github.com/AlgebraicJulia/Catlab.jl) [@patterson2021].

A more formal definition is provided below.

\begin{definition}[Category of $\cat{C}$-sets, $\catSet{C}$]
  For a given schema $\cat{C}$, the \textit{category of $\cat{C}$-sets} is the functor category $\catSet{C} := \Set^{\cat{C}}$, whose objects are functors from $\cat{C}$ to $\Set$ and whose morphisms are natural transformations between those.
\end{definition}

% More formally, for a given schema $\cat{C}$, the \textit{category of $\cat{C}$-sets} is the functor category $\catSet{C} := \Set^{\cat{C}}$, whose objects are functors from $\cat{C}$ to $\Set$ and whose morphisms are natural transformations between those.

In practice, the schema is a category that is finitely presented by generators and relations. As an example, let $\cat{Gr} = \{E \rightrightarrows V\}$ be the category freely generated by two objects, $E$ and $V$, and two parallel morphisms, $\text{src, tgt}: E \rightarrow V$. Then a $\cat{Gr}$-set is a graph, which would be called by graph theorists a ``directed multigraph.'' So, another interpretation of $\cat{C}$-sets is that they are a generalization of graphs to a broad class of combinatorial data structures.

\begin{figure}  
  \begin{center}
  \includegraphics[width=7.5cm]{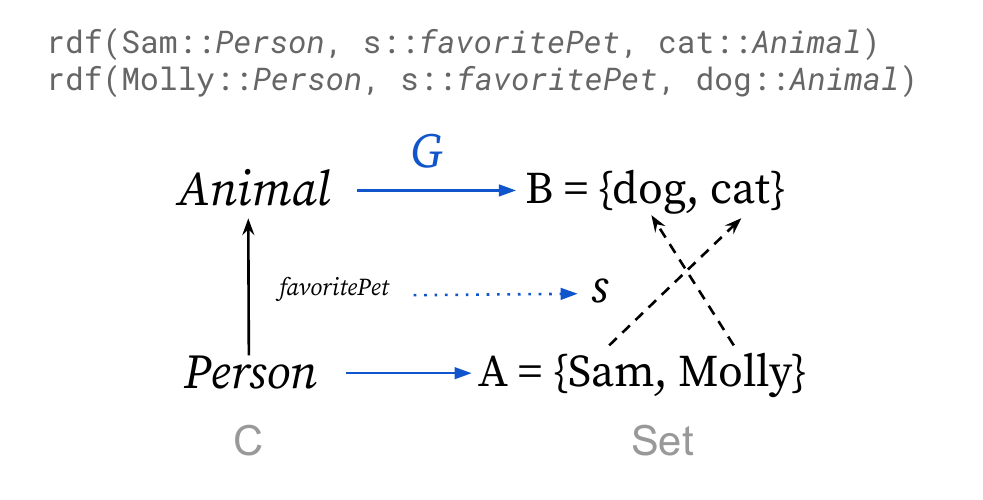}
  \caption{An example $\cat{C}$-set, $G$, that stores data about people's favorite pet. The category of elements contains triples analogous to RDF triples.}
  \label{fig:cset}
  \end{center}
\end{figure}

The \textit{category of elements} \cite{Riehl2016} construction\footnote{The category of elements construction of $X$ may be expressed using $\int X$ in other texts.} of a $\cat{C}$-set $X$ packages the data of $X$ into a category resembling a knowledge graph. Specifically, a morphism in the category of elements can be interpreted as a Resource Description Framework (RDF) triple \cite{Patterson2017}, which is a common text-based serialization format for knowledge and scene graphs. A $\cat{C}$-set is shown in this style in Figure \ref{fig:cset}.

% The category of $\cat{C}$-sets is a topos, an especially well-behaved kind of category in which, for example, all limits and colimits exist.

% \begin{definition}[Attributed $\catSet{C}$]
%   An attributed $\catSet{C}$ is a functor, $F$, from a finitely presented schema category, $\cat{C}$, to $\Set$, where $\cat{C}$ is partitioned using a functor $S:\cat{C} \rightarrow \mathsf{2}$. 
% The preimage $S^{\-1}(0)$ isolates the combinatorial structure.
% The preimage $S^{\-1}(1)$ isolates the attribute structure.
% The preimage $S^{\-1}(0 \rightarrow 1)$ isolates the arrows between the combinatorial structure and the attribute structure.
% \end{definition}

\subsection{Actions as Spans in $\catSet{C}$}
\label{sec:action}

\textit{Action rules} are specified as spans in $\catSet{C}$. Action rules are made up of components similar to those of action schemas in classical representation. Specifically, actions rules are spans ($I \hookleftarrow K \rightarrow O$) in $\catSet{C}$ that consists of the \textit{precondition}, $I$, on the left-hand side, the \textit{effects}, $O$, on the right-hand side, and the \textit{glue}, $K$, in the middle, which gives the data that remains unchanged between the input and the output.

In our approach, we present spans of $\cat{C}$-sets as colimits
of representable functors. A (covariant) representable functor \cite{maclane1971}, $\cat{C}(A,-)$, for a given category $\cat{C}$ and object $A \in \cat{C}$ maps objects, $X \in \cat{C}$, to the set of morphisms that go from from $A$ to $X$. In other words, for a given representable functor pertaining to $A$, objects $X \in \cat{C}$ are mapped to the set of edges that go from $A$ to $X$. A colimit taken across functors like these glues all of the relevant structures together. This conversion ensures that implicit substructure is taken into account when an object, such as $A$, is explicitly identified.
% This conversion is possible because, for every $\cat{C}$-set, $F$, in the action rule, a natural transformation exists from $F$ to a representable functor on $\cat{C}$, as per the Yoneda embedding (see Lemma \ref{def:yoneda} in Appendix \ref{app:constructions}). 
  
% \begin{definition}[Representable Functors]
% A (covariant)\footnote{A representable functor can defined from a covariant ($H^A: \cat{C} \rightarrow \Set$) or a contravariant ($H_A: \cat{C}^{op} \rightarrow \Set$) view \cite{maclane1971}.} representable functor, $H^A: \cat{C} \rightarrow \Set$, is functor that sends:
% \begin{itemize}
%   \item (objects) $x \in \cat{C}, x \mapsto \cat{C}(A, x)$
%   \item (morphisms) $x \xrightarrow{f} y \mapsto \cat{C}(A,x) \xrightarrow{H^A(f)} \cat{C}(A,y)$,
% \end{itemize}
% where $\cat{C}(A,x)$ is the set of all morphisms from $A$ and $x$ in the category $\cat{C}$. $H^A(f)$ postcomposes $f$ with morphisms in $\cat{C}(A,x)$. The functor $H^A$ can also be written as $[\cat{C}, \Set]$.
% \label{def:represent}
% \end{definition}

% Keep in foreground

% Move to appendix. The representables are embedded in the category of c-sets via the yoneda embedding. Lemma or proposition.

The categorical rule specification differs from the classical one in that it takes a declarative approach by not articulating what atoms should be added and removed from the state, but rather discussing what should be in the state and resolving conflicts using the double-pushout (DPO) rewriting procedure which is discussed in a later section.

\subsection{Applicability Using Monomorphisms}

Recall that in classical planning, a precondition is satisfied by a world state when its literals are a subset of the world state or if there exists a logical entailment between the precondition and the world state. In the category-theoretic context, these notions are generalized via monomorphisms.

Monomorphisms generalize the concept of an injective function to arbitrary categories. In $\cat{Set}$, monomorphisms are precisely injective functions. In $\catSet{C}$, monomorphisms are natural transformations such that every component is an injective function, e.g., a monomorphism between graphs is a graph homomorphism such that the vertex and edge maps are both injective. 
More specifically, the monic condition of a monomorphism checks that two entities in the precondition cannot be mapped to the same entity in the world state.

\subsection{Transition Using Double-Pushout (DPO) Rewriting}
\label{sec:dpo}
The action rules are exactly double-pushout (DPO) rewriting rules. Double-pushout (DPO) rewriting is a type of graph rewriting that is particularly well-suited for algebraic approaches to graph transformation. In fact, DPO rewriting generalizes directly from graph rewriting to $\cat{C}$-set rewriting \cite{Brown2022}. The DPO method, described below, is used to compute all possible matches of the preconditions, and to determine which matches are compatible with the effects. The result is a set of transformation steps that can be applied to the target graph. 

DPO rewriting relies on the fundamental concept of a pushout. A pushout is a colimit of a diagram having the shape of a span ($\bullet \leftarrow \bullet \rightarrow \bullet$). Given a span $X \leftarrow Z \rightarrow Y$, a pushout produces an object that is the union of $X$ and $Y$ joined along $Z$, $(X \cup Y) / Z$. A pushout in $\catSet{C}$ is computed by taking the disjoint union of the sets being pushed out, adding the functions between sets based on $\cat{C}$, and quotienting by $Z$.

Pseudocode for the DPO rewriting procedure is given in Algorithm \ref{alg:pushout-complement}. The first step is to find a monomorphism, $m$, that matches $I$ in $X$, as described in the previous subsection. The pushout complement, $f$, is computed provided the morphisms $l$ and $m$. A pushout complement is a map that manages the deletion of entities that form the complement $K / I$. Because $i$ is a monomorphism and we assume the identification and dangling conditions \cite{Brown2022} are met, the pushout complement exists and is unique up to isomorphism. Having constructed the three sides of the square, $l, m, f$, the pushout square can be completed by a unique map $g$. Then, to compute the new world state, the right pushout square is computed provided $f$ and $r$.

$$
% https://q.uiver.app/?q=WzAsNixbMCwwLCJJIl0sWzEsMCwiSyJdLFsyLDAsIk8iXSxbMCwxLCJYIl0sWzEsMSwiWiIsWzI0MCw2MCw2MCwxXV0sWzIsMSwiWSIsWzEyMCw2MCw2MCwxXV0sWzEsMCwiaSIsMix7InN0eWxlIjp7InRhaWwiOnsibmFtZSI6Imhvb2siLCJzaWRlIjoiYm90dG9tIn19fV0sWzEsMiwibyJdLFswLDMsImYiLDIseyJzdHlsZSI6eyJ0YWlsIjp7Im5hbWUiOiJob29rIiwic2lkZSI6InRvcCJ9fX1dLFsxLDQsIlxcdGV4dHtcXGNvbG9ye2xpZ2h0Z3JheX1cXHRleHRjaXJjbGVkezF9fSIsMix7ImNvbG91ciI6WzI0MCw2MCw2MF19LFsyNDAsNjAsNjAsMV1dLFs0LDMsIlxcdGV4dHtcXGNvbG9ye2xpZ2h0Z3JheX1cXHRleHRjaXJjbGVkezJ9fSIsMCx7ImNvbG91ciI6WzMwLDYwLDYwXX0sWzMwLDYwLDYwLDFdXSxbNCw1LCJcXHRleHR7XFxjb2xvcntsaWdodGdyYXl9XFx0ZXh0Y2lyY2xlZHszfX0iLDIseyJjb2xvdXIiOlsxMjAsNjAsNjBdfSxbMTIwLDYwLDYwLDFdXSxbMiw1LCIiLDAseyJjb2xvdXIiOlsxMjAsNjAsNjBdfV0sWzUsMSwiIiwwLHsiY29sb3VyIjpbMTIwLDYwLDYwXSwic3R5bGUiOnsibmFtZSI6ImNvcm5lciJ9fV0sWzMsMSwiIiwwLHsiY29sb3VyIjpbMzAsNjAsNjBdLCJzdHlsZSI6eyJuYW1lIjoiY29ybmVyIn19XV0=
\begin{tikzcd}
  I & K & O \\
  X & \textcolor{gray}{Z} & \textcolor{blue}{Y}
  \arrow["l"', hook', from=1-2, to=1-1]
  \arrow["r", from=1-2, to=1-3]
  \arrow["m"', hook, color={gray}, from=1-1, to=2-1]
  \arrow["f"', color={gray}, from=1-2, to=2-2]
  \arrow["g", color={gray}, from=2-2, to=2-1]
  \arrow[draw={gray}, from=2-2, to=2-3]
  \arrow[draw={gray}, from=1-3, to=2-3]
  \arrow["\lrcorner"{text={gray}, anchor=center, pos=0.125, rotate=180}, draw=none, from=2-3, to=1-2]
  \arrow["\lrcorner"{text={gray}, anchor=center, pos=0.125, rotate=90}, draw=none, from=2-1, to=1-2]
  \end{tikzcd}
$$
% orange: {rgb,255:red,246;green,178;blue,107}
% blue: {rgb,255:red,109;green,158;blue,235}
% green: {rgb,255:red,134;green,222;blue,134}

\begin{algorithm}[t!]
  \caption{Double-Pushout (DPO) Rewriting}
  \label{alg:pushout-complement}
  \begin{algorithmic}
  \Require (action rule) $I \hookleftarrow K \rightarrow O \in \catSet{C}$
  \Require (world) $X \in \catSet{C}$
  \State $m$ = \texttt{FindHomomorphism}($I$,$X$)
  \State $f$ = \texttt{ComputePushoutComplement}($l$, $m$)
  \State $g$ = \texttt{CompletePushout}($l$, $f$, $m$)
  \State $Y$ = \texttt{ComputePushout}($f$, $r$)
  \end{algorithmic}
\end{algorithm}

The time complexity of the \texttt{FindHomomorphism()} subroutine is $O(n^k)$ where $k$ is the size of $I$ and $n$ is the size of the relevant substructure in $X$ which is dictated by the objects in $I$ \cite{Brown2022}. The time complexity of the remaining subroutines is the same as that of computing pushouts in $\cat{Set}$ which is $O(p)$ where $p$ is the sum of the sizes of the sets involved in the span.

\begin{figure}  
  \includegraphics[width=8.5cm]{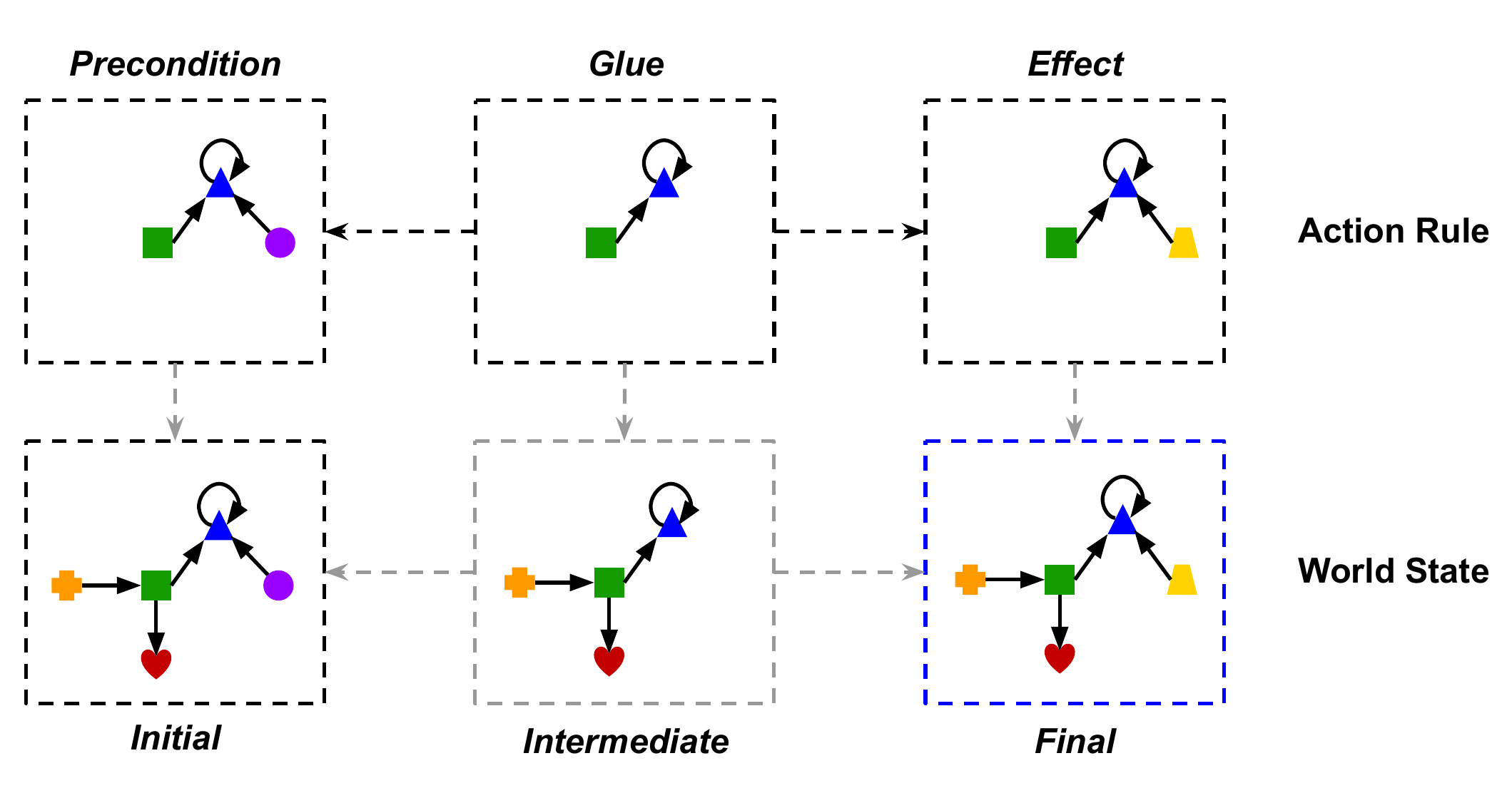}
  \caption{An illustration of how DPO rewriting is executed on a $\cat{C}$-set where $\cat{C}$ defines the shapes and the arrows between them. Each shape in the figure is assigned to a color to help with readability.}
  
  % red: \color{rgb,255:red,196;green,0;blue,0}
  % blue: \color{rgb,255:red,0;green,0;blue,255}
  % green: \color{rgb,255:red,21;green,156;blue,0}
  % orange: \color{rgb,255:red,255;green,153;blue,0}
  % purple: \color{rgb,255:red,153;green,0;blue,255}
  % yellow: \color{rgb,255:red,255;green,211;blue,0}
  
  \label{fig:dpo}
  \end{figure}

As an example, Figure \ref{fig:dpo} shows how an action rule changes the initial state to the final state. In this figure, the rule states that the circle in the initial state should be replaced by a trapezoid and the square and triangle should persist. The initial state, bottom-left, satisfies the precondition because it contains a matching pattern involving the square, triangle, and circle. This means that a monomorphism can be identified from the precondition to the initial state. The intermediate state is the result of identifying a pattern that, when joined with the precondition along the glue, produces the initial state. This removes the circle from the initial state. The final state is then constructed by taking the pushout for the span ($\text{Intermediate} \leftarrow \text{Glue} \rightarrow \text{Effect}$). This produces a pattern where a trapezoid is added in place of the original circle. The example shown in Figure \ref{fig:dpo} demonstrates that rules can involve both the creation and destruction of entities in the world if the precondition contains an entity and the effect does not contain that entity. This allows for a non-monotonicity in updates of the world state.

\section{Comparison}
\label{sec:comparison}

\begin{figure*}[t!]
  \centering
  \includegraphics[width=0.83\textwidth]{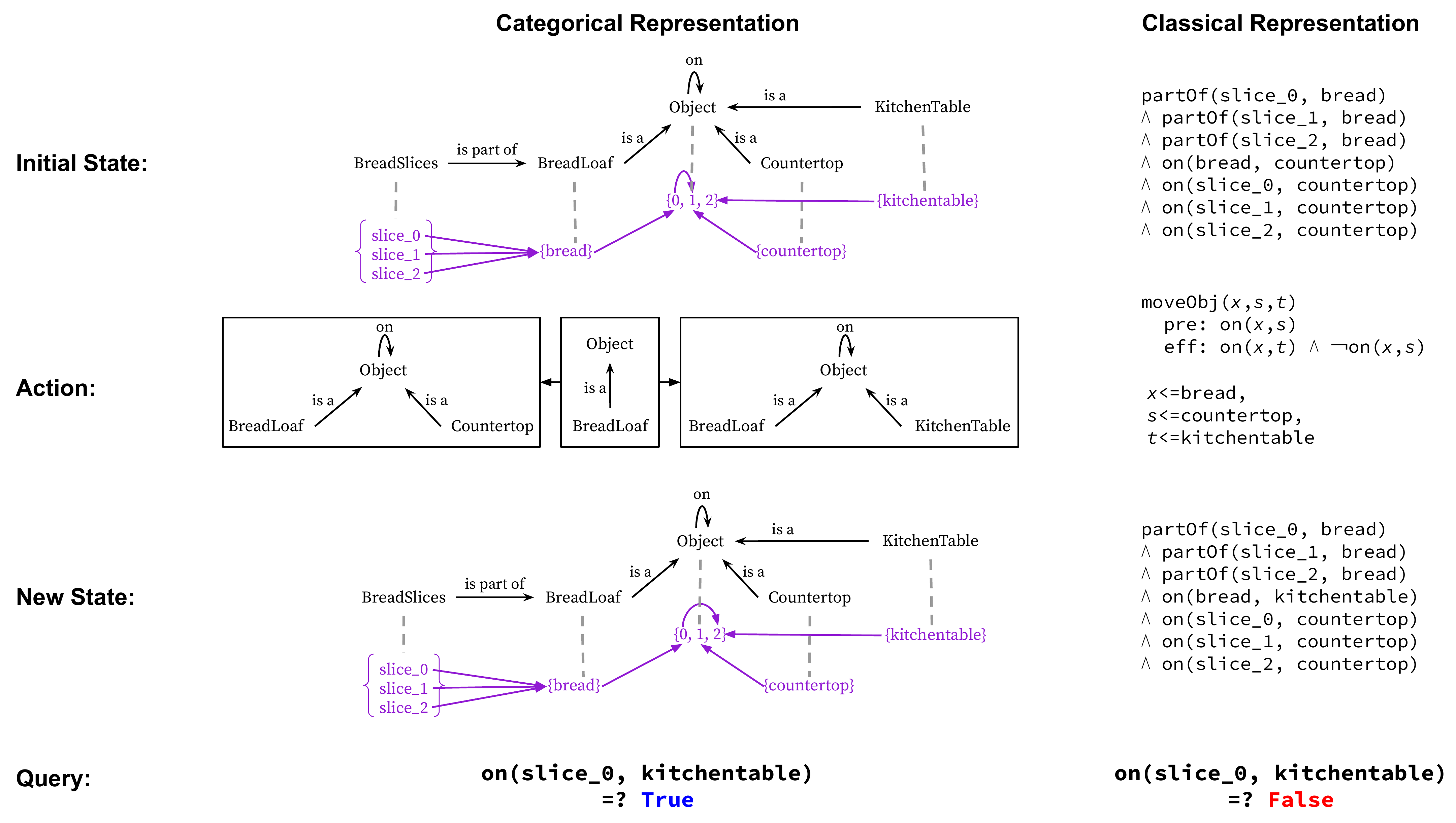}
  \caption{A comparison of states, actions, and inferences made between the categorical representation and the classical representation for an example that moves a loaf of bread from a countertop to a kitchen table. This example illustrates a failure of the classical representation to preserve the global semantics of the world when actions act on only a part of the world. Using the double-pushout method of $\cat{C}$-set rewriting, the categorical representation is able to do so.}
  \label{fig:breadloaf}
\end{figure*}

In this section, we discuss the conceptual differences between the classical and the categorical representations. The comparison is summarized in Table \ref{tab:summary}. We also walk through an example, shown in Figure \ref{fig:breadloaf}, that highlights the limitations of the classical representation in tracking implicit effects. In this example domain, a bread loaf (\texttt{bread loaf}) and slices of the loaf (\texttt{slice\_0}, \texttt{slice\_1}, \texttt{slice\_2}) are on a countertop (\texttt{countertop}). The goal is to move the bread loaf, and implicitly, all its slices, from the countertop to the kitchen table (\texttt{kitchentable}).

\begin{table*}[t!]
  \begin{center}
    \begin{tabular}{l p{2in} p{2in}}
     \toprule
     \textbf{Representation} & \textbf{Categorical} & \textbf{Classical} \\ 
     \midrule
     State, $S$ & $\cat{C}$-set & Conjunction of literals\\
     Action, $A$ & Spans in $\catSet{C}$ containing: Preconditions, Glue, Effects & Action model containing: Parameters, Preconditions, Effects\\
     Applicability & Monomorphisms in $\catSet{C}$ & Subset inclusion \\
     Transition, $\gamma$ & DPO rewriting & Set-based addition and subtraction \\
     \bottomrule
  \end{tabular}
  \end{center}
  \caption{A summary of the differences between the classical representation and the categorical representation aligned to the state transition system model for planning.}
  \label{tab:summary}
  \end{table*}

\subsection{Handling Structured Knowledge}

This example presents a few noteworthy semantic features. A planning representation in this domain must be able to encode the following facts, explicitly or implicitly, at different points in time:

  \begin{enumerate}[label=(\alph*),leftmargin=2\parindent]
    \item the bread slices are part of the bread loaf
    \item the bread loaf is on the countertop
    \item the bread slices are on the countertop
    \item the bread loaf is on the kitchen table
    \item the bread slices are on the kitchen table
  \end{enumerate}

In the case of the categorical representation, fact (a) is captured by the morphism $\mathrm{is\ part\ of}: \mathrm{BreadSlices} \rightarrow \mathrm{BreadLoaf}$ in the schema. Fact (b) about the bread loaf being on the countertop is reified through an abstract type called $\mathrm{Object}$. This is done so that the morphism $\mathrm{on} : \mathrm{Object} \to \mathrm{Object}$ can represent the general notion of an item being on top of another item, instead of the more specific relation of a bread loaf being on a countertop. Fact (c) is then captured by the composite morphism $\mathrm{on} \circ \mathrm{is\ a} \circ \mathrm{is\ part\ of}: \mathrm{BreadSlices} \rightarrow \mathrm{Object}$ \footnote{The operator $\circ$ represents a composite function which can be read as "$f$ following $g$" or $(f \circ g)(x) = f(g(x))$.}. In the classical representation, fact (a) is captured by the propositions \texttt{partOf(slice\_\textit{n}, bread)} for $n = 0,1,2$. Fact (b) and (d) are captured by the proposition \texttt{on(bread, countertop)} and \texttt{on(bread, kitchentable)}. Fact (c) and (e) are captured by the propositions \texttt{on(slice\_\textit{n}, countertop)} and \texttt{on(slice\_\textit{n}, kitchentable)}. Intuitively, because the bread loaf is on the countertop, the slices that make up the loaf are also on the countertop. In the categorical representation, this is captured using a composite morphism. In the classical representation, this is done by explicitly stating for each slice, that it is on the countertop.

% CT --> FOL
% Translate objects in El(F) to typed constants
% Translate relations in El(F) to typed predicates
% Deciding applicability of rules combinatorially vs deciding applicability of rules using theorem proving (monic morphism)
% Logical updates is done combinatorially vs. logical updates done using resolution and unification (frame problem)

\subsection{Handling Applicability of Actions}

In the categorical representation, applicability of an action is determined by the existence of a monomorphism in $\catSet{C}$ from the rule input to the world state. Recall that a representable functor maps an object, $x \in \cat{C}$, to the set of morphisms that involve $x$. When you present an action using colimits of representables, there are both explicit conditions given by the representables and implicit conditions that appear when the representable is computed. This provides a mechanism for having implicit conditions represented in the action schemas. In this example, the initial state satisfies the input of the action rule because it contains $\mathrm{BreadLoaf}$, $\mathrm{Object}$, and $\mathrm{Countertop}$ and its related morphisms. In the classical representation, applicability of the action \texttt{moveObj(bread, countertop, kitchentable)} is determined by whether or not the world state contains the element \texttt{on(bread, countertop)} in the set of propositions. 

\subsection{Handling Implicit Effects}

Now that we have determined that this action is applicable in both representations, we can apply the action. In the categorical representation, this action is modeled by a span in $\catSet{C}$  whose left foot includes knowledge about the bread loaf being on the countertop and whose right foot includes knowledge about the bread loaf being on the kitchen table. The apex of the span states that the bread loaf itself is preserved throughout this change. In the classical representation, the action schema describes the generic action of moving items \texttt{x} from one location, \texttt{s}, to another, \texttt{t}. The action operator is grounded by assigning \texttt{x} to the \texttt{breadloaf}, \texttt{s} to the \texttt{countertop}, and \texttt{t} to the \texttt{kitchentable}. Once this action is applied, a desirable outcome would be for the new state to account for the movement of the bread slices from the countertop to the kitchen table because they are part of the breadloaf. In the categorical representation, the same composite morphism that existed in the initial state exists in the final state; however, the target of the morphism has changed from the $\mathrm{countertop}$ to the $\mathrm{kitchentable}$. This captures the implicit change that occurred to the bread slice locations. In the classical representation, the new state captures the fact that the breadloaf is on the kitchen table, but does not capture the fact that the bread slices are on the kitchen table. This error would likely cause a planner to instruct an agent to move each bread slice individually due to an inconsistency in the world state.

\section{Conclusion}
\label{sec:conclusion}

The limitations of classical planning representation languages in tracking implicit preconditions and effects have motivated us to propose an alternative world state representation based on the category-theoretic concepts of $\cat{C}$-sets and DPO rewriting. Our categorical representation accommodates structured knowledge about the world state and formalizes a model of the application domain using a user-provided ontology. This method provides formal semantics for using knowledge graphs and relational databases to model world states and updates in planning. Our comparison between the classical and categorical planning representation languages demonstrates that our proposed representation is more structured and has advantages over the classical one in handling complex planning scenarios. Future work include designing mechanisms for transferring plans between domains and extending this system to handle non-combinatorial data, such as numeric data. We believe that our proposed representation has the potential to significantly enhance the effectiveness and efficiency of planning systems in robotic task planning domains.

% This makes it possible to do planning using scene graphs. Our representation formalizes the grounding and lifting relationship between atoms that preserves semantics and handles the frame problem. 

\section*{Acknowledgments}

This work was partially funded by the U.S. Defense Advanced Research Projects Agency under contract \#HR00112220004. Any opinions, findings, and conclusions or recommendations expressed in this paper are those of the authors and do not necessarily reflect these agencies' views. We thank Kristopher Brown and Owen Lynch from the Topos Institute for heavily contributing to the development of DPO rewriting and $\cat{C}\text{-sets}$ in AlgebraicJulia. We also thank the students of the Ruiz HCI Lab---Alexander Barquero, Rodrigo Calvo, Niriksha Regmi, Daniel Delgado, Andrew Tompkins---who built the task guidance platform that inspired the examples in this paper. Lastly, we thank Dana Nau for valuable feedback about this paper.

\bibliography{ref}

\begin{thebibliography}{29}
\providecommand{\natexlab}[1]{#1}

\bibitem[{Abramsky and Coecke(2008)}]{Abramsky2008}
Abramsky, S.; and Coecke, B. 2008.
\newblock {Categorical Quantum Mechanics}.
\newblock \emph{Handbook of Quantum Logic and Quantum Structures vol II},
  1--63.

\bibitem[{Agia et~al.(2021)Agia, Jatavallabhula, Khodeir, Miksik, Vineet,
  Mukadam, Paull, and Shkurti}]{Agia2022}
Agia, C.; Jatavallabhula, K.~M.; Khodeir, M.; Miksik, O.; Vineet, V.; Mukadam,
  M.; Paull, L.; and Shkurti, F. 2021.
\newblock {TASKOGRAPHY: Evaluating robot task planning over large 3D scene
  graphs}.
\newblock \emph{5th Conference on Robot Learning}, 1--18.

\bibitem[{Aguinaldo et~al.(2021)Aguinaldo, Bunker, Pollard, Shukla, Canedo,
  Quiros, and Regli}]{Aguinaldo2021}
Aguinaldo, A.; Bunker, J.; Pollard, B.; Shukla, A.; Canedo, A.; Quiros, G.; and
  Regli, W. 2021.
\newblock {RoboCat: A Category Theoretic Framework for Robotic Interoperability
  Using Goal-Oriented Programming}.
\newblock \emph{IEEE Transactions on Automation Science and Engineering}, 1--9.

\bibitem[{Aguinaldo and Regli(2021)}]{Aguinaldo2021a}
Aguinaldo, A.; and Regli, W. 2021.
\newblock {A Graphical Model-Based Representation for PDDL Plans using Category
  Theory}.
\newblock In \emph{ICAPS 2021 Workshop XAIP}.

\bibitem[{Amiri, Chandan, and Zhang(2022)}]{Amiri2022}
Amiri, S.; Chandan, K.; and Zhang, S. 2022.
\newblock {Reasoning with Scene Graphs for Robot Planning under Partial
  Observability}.
\newblock \emph{IEEE Robotics and Automation Letters}, 7(2).

\bibitem[{Baez and Stay(2011)}]{Baez2011}
Baez, J.; and Stay, M. 2011.
\newblock {Physics, topology, logic and computation: A Rosetta Stone}.
\newblock \emph{Lecture Notes in Physics}, 813: 95--172.

\bibitem[{Beetz et~al.(2018)Beetz, Bessler, Haidu, Pomarlan, Bozcuoglu, and
  Bartels}]{Beetz2018}
Beetz, M.; Bessler, D.; Haidu, A.; Pomarlan, M.; Bozcuoglu, A.~K.; and Bartels,
  G. 2018.
\newblock {Know Rob 2.0 - A 2nd Generation Knowledge Processing Framework for
  Cognition-Enabled Robotic Agents}.
\newblock \emph{Proceedings - IEEE International Conference on Robotics and
  Automation}, 512--519.

\bibitem[{Breiner, Pollard, and Subrahmanian(2019)}]{Breiner2019a}
Breiner, S.; Pollard, B.; and Subrahmanian, E. 2019.
\newblock {Functorial Model Management}.
\newblock \emph{Proceedings of the Design Society: International Conference on
  Engineering Design}, 1(1): 1963--1972.

\bibitem[{Breiner, Subrahmanian, and Sriram(2016)}]{Breiner2016}
Breiner, S.; Subrahmanian, E.; and Sriram, R.~D. 2016.
\newblock {Modeling the Internet of Things: A foundational approach}.
\newblock \emph{ACM International Conference Proceeding Series}, Part F1271:
  38--41.

\bibitem[{Brown et~al.(2021)Brown, Patterson, Hanks, and Fairbanks}]{Brown2022}
Brown, K.; Patterson, E.; Hanks, T.; and Fairbanks, J. 2021.
\newblock {Computational category-theoretic rewriting}.
\newblock \emph{Lecture Notes in Computer Science (including subseries Lecture
  Notes in Artificial Intelligence and Lecture Notes in Bioinformatics)}.

\bibitem[{Censi(2016)}]{Censi2015}
Censi, A. 2016.
\newblock A Mathematical Theory of Co-Design.
\newblock Technical report, Laboratory for Information and Decision Systems,
  MIT.

\bibitem[{Chang et~al.(2023)Chang, Ren, Xu, Li, Chen, and
  Hauptmann}]{Chang2023}
Chang, X.; Ren, P.; Xu, P.; Li, Z.; Chen, X.; and Hauptmann, A. 2023.
\newblock {A Comprehensive Survey of Scene Graphs: Generation and Application}.
\newblock \emph{IEEE Transactions on Pattern Analysis and Machine
  Intelligence}, 45(1): 1--26.

\bibitem[{Galindo et~al.(2008)Galindo, Fern{\'{a}}ndez-Madrigal,
  Gonz{\'{a}}lez, and Saffiotti}]{Galindo2008}
Galindo, C.; Fern{\'{a}}ndez-Madrigal, J.~A.; Gonz{\'{a}}lez, J.; and
  Saffiotti, A. 2008.
\newblock {Robot task planning using semantic maps}.
\newblock \emph{Robotics and Autonomous Systems}, 56(11): 955--966.

\bibitem[{Ghallab, Nau, and Traverso(2004)}]{Ghallab2004}
Ghallab, M.; Nau, D.; and Traverso, P. 2004.
\newblock \emph{{Automated Planning: Theory and Practice}}.
\newblock Morgan Kaufmann.
\newblock ISBN 9781558608566.

\bibitem[{Gil(1990)}]{Gil1990}
Gil, Y. 1990.
\newblock {Description Logics and Planning}.
\newblock \emph{AI Magazine}, 26(2): 1.

\bibitem[{Kang and Choi(2009)}]{Kang2009}
Kang, D.; and Choi, H.-j. 2009.
\newblock {Hierarchical Planning through Operator and World Abstraction using
  ontology for home service robots}.
\newblock \emph{2009 11th International Conference on Advanced Communication
  Technology}, 2209--2214.

\bibitem[{Knoblock et~al.(1998)Knoblock, Barrett, Christianson, Friedman, Kwok,
  Golden, Penberthy, Smith, Sun, and Weld}]{Knoblock1998}
Knoblock, C.; Barrett, A.; Christianson, D.; Friedman, M.; Kwok, C.; Golden,
  K.; Penberthy, S.; Smith, D.~E.; Sun, Y.; and Weld, D. 1998.
\newblock {PDDL | The Planning Domain Definition Language}.
\newblock Technical report, Yale Center for Computational Vision and Control.

\bibitem[{Leinster(2016)}]{Leinster2016}
Leinster, T. 2016.
\newblock \emph{{Basic Category Theory}}.
\newblock Cambridge University Press.

\bibitem[{MacLane(1971)}]{maclane1971}
MacLane, S. 1971.
\newblock \emph{Categories for the Working Mathematician}.
\newblock New York: Springer-Verlag.
\newblock Graduate Texts in Mathematics, Vol. 5.

\bibitem[{Master et~al.(2020)Master, Patterson, Yousfi, and
  Canedo}]{Master2019}
Master, J.; Patterson, E.; Yousfi, S.; and Canedo, A. 2020.
\newblock {String Diagrams for Assembly Planning}.
\newblock \emph{Lecture Notes in Computer Science (including subseries Lecture
  Notes in Artificial Intelligence and Lecture Notes in Bioinformatics)}.

\bibitem[{Miao, Jia, and Sun(2023)}]{Miao2023}
Miao, R.; Jia, Q.; and Sun, F. 2023.
\newblock {Long-term robot manipulation task planning with scene graph and
  semantic knowledge}.
\newblock \emph{Robotic Intelligence and Automation}.

\bibitem[{Patterson(2017)}]{Patterson2017}
Patterson, E. 2017.
\newblock Knowledge Representation in Bicategories of Relations.
\newblock \emph{CoRR}, abs/1706.00526.

\bibitem[{Patterson, Lynch, and Fairbanks(2021)}]{Patterson2021}
Patterson, E.; Lynch, O.; and Fairbanks, J. 2021.
\newblock {Categorical Data Structures for Technical Computing}.
\newblock \emph{Compositionality}, 4(5): 1--27.

\bibitem[{Riehl(2016)}]{Riehl2016}
Riehl, E. 2016.
\newblock \emph{{Category theory in context}}.
\newblock ISBN 9780486809038.

\bibitem[{Silver et~al.(2021)Silver, Chitnis, Curtis, Tenenbaum,
  Lozano-P{\'{e}}rez, and Kaelbling}]{Silver2021}
Silver, T.; Chitnis, R.; Curtis, A.; Tenenbaum, J.; Lozano-P{\'{e}}rez, T.; and
  Kaelbling, L.~P. 2021.
\newblock {Planning with Learned Object Importance in Large Problem Instances
  using Graph Neural Networks}.
\newblock \emph{35th AAAI Conference on Artificial Intelligence, AAAI 2021},
  13B: 11962--11971.

\bibitem[{Spivak and Kent(2011)}]{Spivak2011}
Spivak, D.~I.; and Kent, R.~E. 2011.
\newblock {Ologs: a categorical framework for knowledge representation}.
\newblock \emph{PLOS ONE}.

\bibitem[{Tenorth and Beetz(2017)}]{Tenorth2017}
Tenorth, M.; and Beetz, M. 2017.
\newblock {Representations for robot knowledge in the KNOWROB framework}.
\newblock \emph{Artificial Intelligence}, 247: 151--169.

\bibitem[{Thi{\'{e}}baux, Hoffmann, and Nebel(2005)}]{Thiebaux2005}
Thi{\'{e}}baux, S.; Hoffmann, J.; and Nebel, B. 2005.
\newblock {In defense of PDDL axioms}.
\newblock \emph{Artificial Intelligence}, 168(1-2): 38--69.

\bibitem[{Wilkins and DesJardins(2001)}]{Wilkins2001}
Wilkins, D.~E.; and DesJardins, M. 2001.
\newblock {A call for knowledge-based planning}.
\newblock \emph{AI Magazine}, 22(1): 99--115.

\end{thebibliography}

\end{document}